\documentclass{article}
\usepackage{times}
\usepackage{epsfig}
\usepackage{graphicx}
\usepackage{amsmath}
\usepackage{amssymb}
\usepackage{multirow}
\usepackage{algorithm}
\usepackage{dsfont}
\usepackage{bbm}
\usepackage{array}
\usepackage{comment}
\usepackage{pifont}
\usepackage{booktabs}

\usepackage[accepted]{icml}

\icmltitlerunning{Deep Contextual Recurrent Residual Networks for Scene Labeling}

\begin{document} 

\twocolumn[
\icmltitle{Deep Contextual Recurrent Residual Networks for Scene Labeling}

\begin{icmlauthorlist}
\icmlauthor{T. Hoang Ngan Le}{cmu}
\icmlauthor{Chi Nhan Duong}{cmu,con}
\icmlauthor{Ligong Han}{cmu}
\icmlauthor{Khoa Luu}{cmu}
\icmlauthor{Marios Savvides}{cmu}
\icmlauthor{Dipan Pal }{cmu} 
\end{icmlauthorlist}

\icmlcorrespondingauthor{T. Hoang Ngan Le} {thihoanl@andrew.cmu.edu}
\icmlaffiliation{cmu}{CyLab Biometrics Center and the Department of Electrical and Computer Engineering,Carnegie Mellon University, Pittsburgh, PA, USA.\\}
\icmlaffiliation{con}{Computer Science and Software Engineering, Concordia University, Montreal, Quebec, Canada.\\}

\vskip 0.3in
]

\printAffiliationsAndNotice{}

\begin{abstract}
Designed as extremely deep architectures, deep residual networks which provide a rich visual representation and offer robust convergence behaviors have recently achieved exceptional performance in numerous computer vision problems. Being directly applied to a scene labeling problem, however, they were limited to capture long-range contextual dependence, which is a critical aspect. To address this issue, we propose a novel approach, \textbf{Contextual Recurrent Residual Networks (CRRN)} which is able to simultaneously handle rich visual representation learning and long-range context modeling within a fully end-to-end deep network. Furthermore, our proposed end-to-end CRRN is completely trained from scratch, without using any pre-trained models in contrast to most existing methods usually fine-tuned from the state-of-the-art pre-trained models, e.g. VGG-16, ResNet, etc. The experiments are conducted on four challenging scene labeling datasets, i.e. SiftFlow, CamVid, Stanford background and SUN datasets, and compared against various state-of-the-art scene labeling methods. 
\end{abstract}

\section{Introduction}

\begin{figure}[t]
	\centering \includegraphics[width=8.3cm]{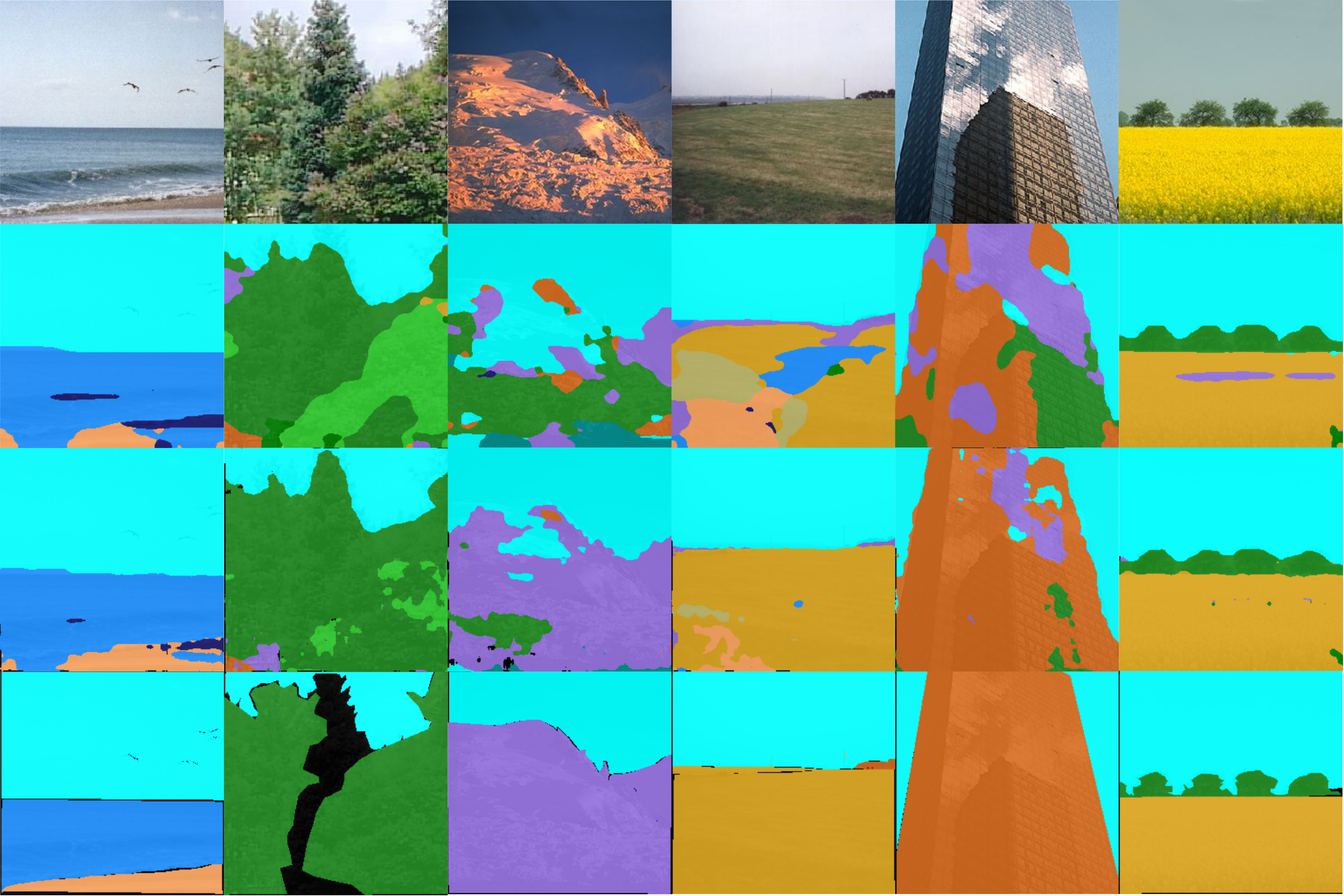}
	\caption{Examples of scene labeling results. From top to bottom: the input images, the segmentation results from \cite{shuai2015dag}, our CRRN segmentation results, and the ground truth.}
	\label{DAGResNet_examples}
\end{figure}

Scene labeling has played an important role in many applications in computer vision and machine learning. This problem is known as semantic segmentation and refers to associating one class to each pixel in a scene image. 
To address this issue, a large body of researches have recently proposed different approaches mainly focusing on contextual information via recurrent neural network \cite{shuai2015integrating, fan2016multi, yang2014context, shotton2006textonboost} or enriching visual representations via convolutional neural network \cite{farabet2013learning, long2015fully, pinheiro2014recurrent, zheng2015conditional}. 
However, scene labeling problem in the real world needs both information of the context dependencies and visual representation. For example, powerful visual representation is capable to discriminate a road from a beach; but it may not effective enough to tell a patch of sand belongs to the side of a road or to a beach. In such circumstance, the context presented in the whole scene can show its advantage to describe them.
Indeed, the roles of contextual information and powerful descriptive visual representation are equally important in the scene labeling problem. 

To effectively address the scene labeling problem, we propose a novel deep network named \textbf{Contextual Recurrent  Residual Network (CRRN)} that inherits all the merits of sequence learning information and residual learning in order to simultaneously model \textit{long-range contextual information} and learn \textit{powerful visual representation} within a \textit{single deep network}. Our proposed CRRN deep network consists of three parts corresponding to sequential input data, sequential output data and hidden state. Each unit in hidden state is designed as a combination of two components: a context-based component via sequence learning and a visual-based component via residual learning. That means, each hidden unit in our proposed CRRN simultaneously (1) learns long-range contextual dependencies via context-based component. The relationship between the current unit and the previous units is performed as sequential information under an undirected cyclic graph (UCG) and (2) provides powerful encoded visual representation via residual component which contains blocks of convolution and/or batch normalization layers equipped with an identity skip connection. Furthermore, unlike previous scene labeling approaches \cite{shuai2015integrating, fan2016multi, byeon2015scene}, our method is not only able to exploit the long-range context and visual representation but also formed under a fully-end-to-end trainable system that effectively leads to the optimal model. In contrast to other existing deep learning network which are based on pre-trained models, our fully-end-to-end CRRN is completely trained from scratch.

\section{Related Work}
\label{sec:relatedwork}

Scene labeling is arguably one of the hardest challenges in computer vision. It requires the algorithms to have much more finesse than those that are only required to tackle image scale object recognition for instance. Nonetheless, a lot of studies have focused on this challenging problem in the past and have made considerable progress recently. Generally, scene labeling  methods can be divided into three categories as follows.

\textbf{Graphical model approaches:} In the past, using traditional vision techniques, scene labeling was approached from a undirected graphical model paradigm utilizing markov random fields (MRF) and conditional random fields (CRF) \cite{Liu2016, Lazebnik_ECCV10, yang2014context}. In \cite{Liu2016}, SIFT flow was utilized as features, whereas as  \cite{Lazebnik_ECCV10} utilized $k$-nearest neighbors in a retrieval dataset to classify superpixels. One of the first studies to utilize contextual information within the MRF framework was \cite{yang2014context}. Similar efforts have been made for using CRFs on unary and pairwise image features \cite{shotton2006textonboost}. Parametric and non-parametric techniques were also combined to model global order dependencies to provide more information and context \cite{shuai2015integrating}. Higher order dependencies were also modeled using a fully connected graph \cite{zhang2012efficient, roy2014scene}.

\textbf{ConvNet-based approaches:} In recent years, deep learning techniques have started to become ubiquitous in scene labeling. One of the first studies to apply convolutional neural networks (deep CNNs) to scene labeling was \cite{farabet2013learning}, which stacked encompassing windows from different scales to serve as context.  This inspired other studies in which fully convolutional networks were used instead \cite{long2015fully} utilizing higher model complexity. Both these techniques used filter based models to incorporate context. Recently, recurrent models have started to gain popularity. For example  \cite{pinheiro2014recurrent}, where the image is passed through a CNN multiple times in sequence \emph{i.e.} the output of the CNN is fed into the same CNN again. As an interesting study, \cite{zheng2015conditional} modeled a CRF as a neural network that is applied iteratively to an input, thereby qualifying as a recurrent model. Inference is done through convergence of the neural network output to a fixed point. Recently, deep residual networks (ResNets) \cite{he2015deep} have emerged as a family of extremely deep architectures showing compelling accuracy and desirable convergence behaviors. They consist of blocks of convolutional and/or batch normalization layers equipped with an identity skip connection. The identity connection helps to address the vanishing gradient problem and allows the ResNets to robustly train using standard stochastic gradient descent despite very high model complexity. This enables ResNets to extract very rich representations of images that perform exceedingly well in image recognition and object detection challenges \cite{he2016identity}. The extremely deep architectures in ResNets show compelling accuracy and robust convergence behaviors and achieve state-of-the-art performance on many challenging computer vision tasks on ImageNet \cite{ImageNet_db}, PASCAL Visual Object Classes (VOC) Challenge \cite{Everingham15} and Microsoft Common Objects in Context (MS COCO) \cite{COCO_db} competitions. Nonetheless, they are feed forward models that do not explicitly encode contextual information and typically cannot be applied to sequence modeling problems. On the other hand, they are able to \textit{ effectively learn visual representations but limited to model long-range context explicitly}

\textbf{Recurrent-based approaches:} In recent years, vision data is being interpreted as sequences leading to the successful application of RNNs (and their variants, e.g. Long-Short Term Memories (LSTMs), Gated Recurrent Units (GRUs), etc.,) to vision problems. For instance, \cite{zuo2015convolutional} and \cite{graves2012offline} applied 1-D RNNs and multi-dimensional RNNs to model contextual dependencies in object recognition/image classification and offline handwriting recognition respectively. 2-D LSTMs instead were applied to scene parsing \cite{byeon2015scene}, Lately, scene labeling \cite{shuai2015integrating}, object segmentation \cite{VisinKCBMC15}, have been reformulated as sequence learning, thereby allowing RNNs to be applied directly. Scene labeling, in particular, has seen the use of RNNs coupled with Directed Acyclic Graphs (DAGs) to model an image as a sequence \cite{shuai2015integrating, fan2016multi, byeon2015scene}. There have also been a few studies that utilize RNNs to compute visual representations \cite{mnih2014recurrent, lrcn2014}. Clearly, RNN-based approaches are \textit{ effective in context modeling but lack the ability to learn visual representation}.

\textbf{Drawbacks of the current approaches are:} (1) The ConvNet-based approaches model makes use of convolutional filters which allow them to learn the short-range context of surrounding neighbors designed by these filers. Therefore they are limited to generalize to long-range contexts dependencies. (2) The RNN-based approaches usually utilize a feature extractor that is independent of the sequence modeling framework, in many cases being trained component wise and not end-to-end \cite{shuai2015dag}. (3) Purely RNN based approaches fail to extract robust visual features during sequence learning itself. This is due to simple linear models being used as the recurrent internal models. 

\begin{figure}[!t]
	\centering \includegraphics[width=1.0\columnwidth]{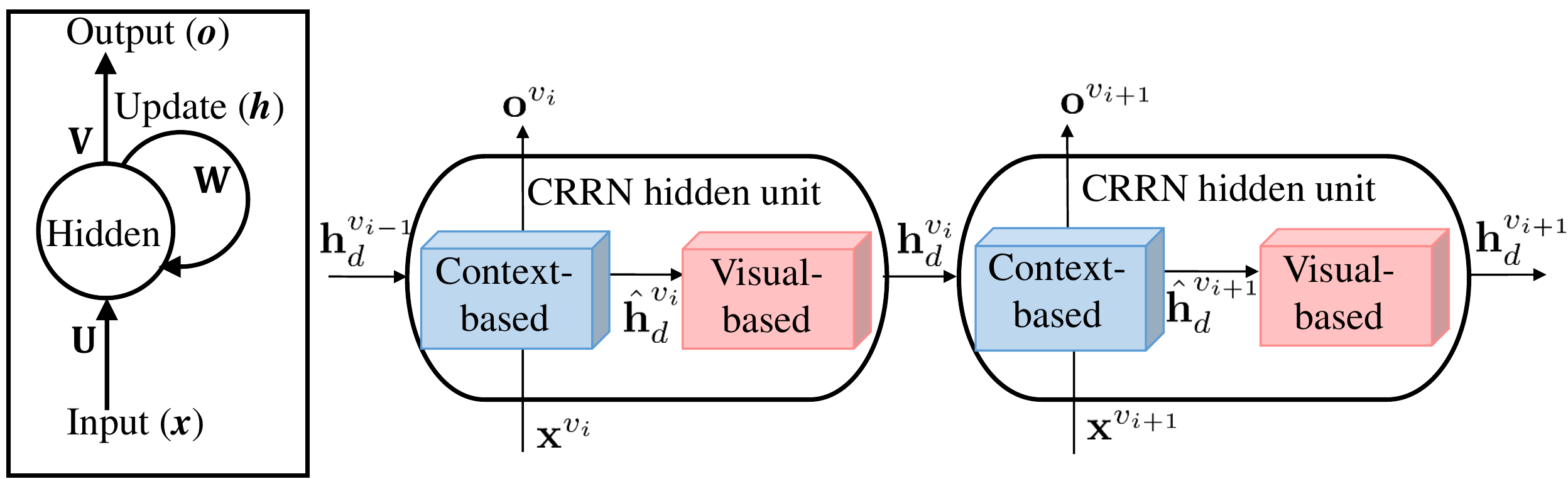}
		\caption{The proposed CRRN and the unfolding in time of the computation involved in its forward procedure.}
	\label{fig:CRRN_folded_unfolded}
\end{figure}

\section{Proposed Contextual Recurrent Residual Networks (CRRN)}
This section presents our proposed Contextual Recurrent Residual Networks (CRRN) for scene labeling problem. The proposed CRRN architecture is first described in Sec.\ref{subsec:CRRN_architecture}. Then, the model learning and the inference are detailed in Sec. \ref{sec:Model_Learning} and Sec. \ref{subsec:CRRN_inference}, respectively.

\subsection{The Proposed Network Architecture}
\label{subsec:CRRN_architecture}

Our proposed CRRN is designed as a composition of multiple CRRN units. 
Each of them consists of input, hidden, and output units. 
In the core of each CRRN hidden unit, two components, i.e. context-based and visual-based components, are employed  to simultaneously handle two significant tasks.
The former helps to handle the contextual knowledge embedding process while the latter tries to increase the robustness of visual representation extracted by the model.
With this structure, one component can benefit from the other and provide more robust output as a result. On one hand, the powerful visual representation from the visual-based component allows the model to have better descriptors for each local patch and results in the better contextual knowledge embeddings. On the other hand, with better memory contextual dependencies from context-based component, highly discriminative descriptors can be extracted. 
Fig. \ref{fig:CRRN_folded_unfolded} demonstrates the folding CRRN on the left and unfolding CRRN in time on the right. 
\begin{figure}[!t]
	\centering \includegraphics[width=1.0\columnwidth]{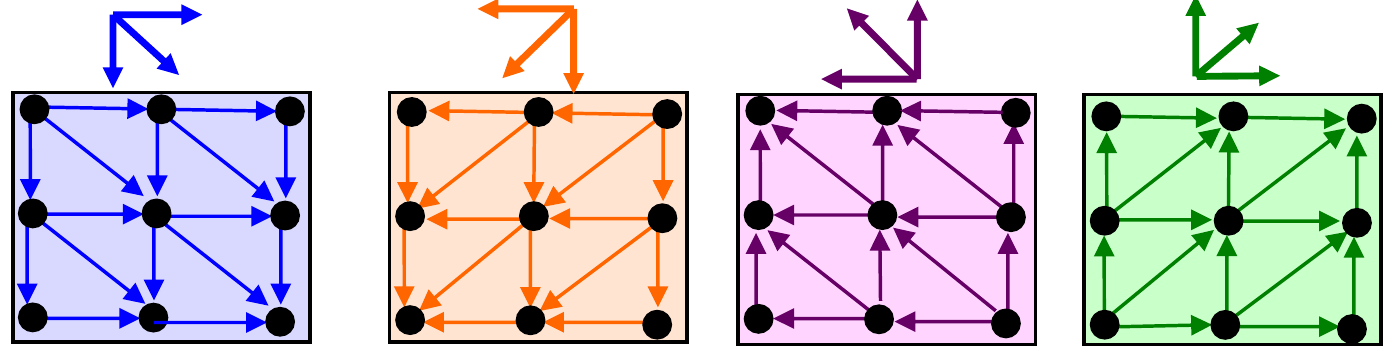}
		\caption{ Decomposition of UCG into four DAGs. From left to right: southeast, southwest, northwest and northeast, where $\bullet$ denotes as a vertex}
	\label{fig:decompose}
\end{figure}
\label{subsec:forward}
\begin{figure}[!t]
	\centering \includegraphics[width=1.0\columnwidth]{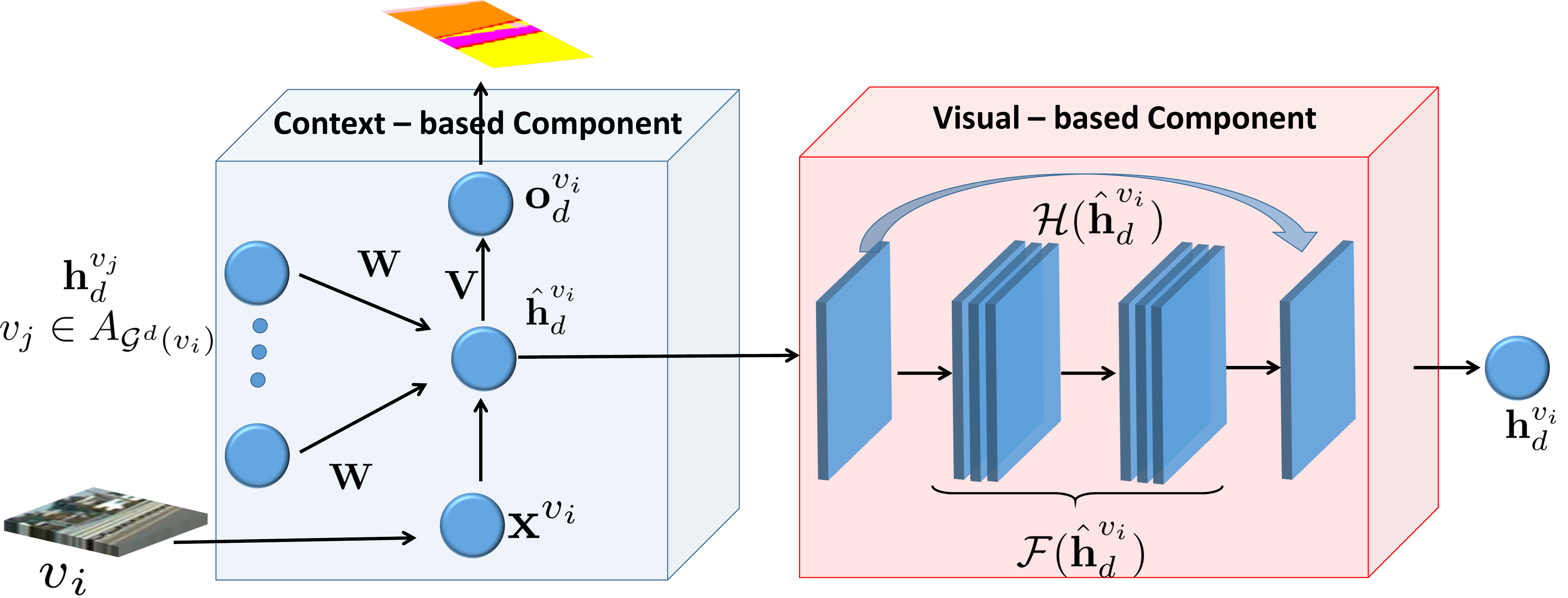}
	\caption{The forward procedure of CRRN at vertex $\textbf{v}_i$ of one CRRN unit with two components corresponding to context information modeling and visual representation learning}
	\label{fig:forward}
\end{figure}

Moreover, unlike previous approaches that are only able to capture the short-range context presented in small local input patches, our CRRN aims at modeling larger-range context by utilizing a graph structure over an image. As a result, the long-range contextual dependencies among different regions of an image can be efficiently modeled and, therefore, increasing the capability of the entire model.
In particular, given an input image, it is first divided into $N$ non-overlapping blocks and their interactions are represented as an undirected cyclic graph (UCG). However, an UCG is unable to unroll as the forward-backward style deep model \cite{shuai2015dag, fan2016multi} due to its loopy structures. 
Therefore, to address this issue, we first decompose the UCG into four directed acyclic graphs (DAGs) along southeast, southwest, northwest and northeast directions as given in Fig. \ref{fig:decompose}. Then the contextual dependencies presented in each DAG are modeled by our CRRN. Finally, these information is combined to produce the final prediction.

\begin{figure*}[t]
	\centering \includegraphics[width=1.8\columnwidth]{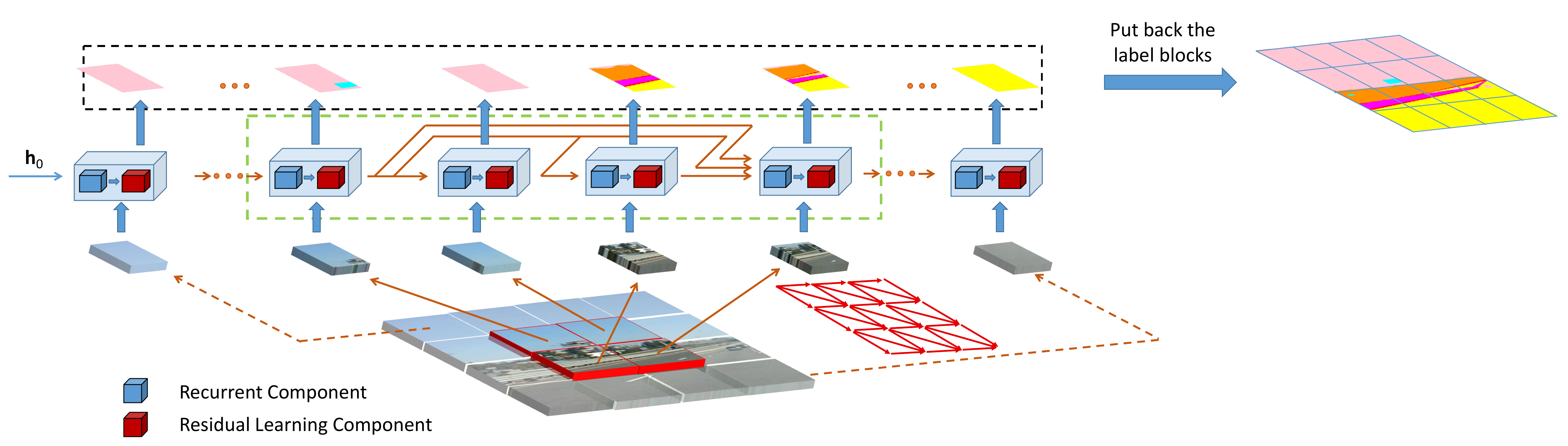}
	\caption{An illustration of our proposed CRRN architecture at one direction (southeast) }
	\label{fig:CRRN_architecture}
\end{figure*}

Formally, an input image $\textbf{I}$ is first divided into $N$ non-overlapping blocks $\{v_i\}_{i=1,2,.., N}$. Each block is then considered as a vertex in four DAGs $\mathcal{G}^1$, $\mathcal{G}^2$, $\mathcal{G}^3$, $\mathcal{G}^4$ corresponding to the four directions. Each DAG is formed as $\mathcal{G}^d = \{\mathcal{V,E}\}|_{d=1}^{4}$, where $\mathcal{V}=\{v_i\}_{i=1,2,.., N}$, and $\mathcal{E}=\{e_{ij}\}$ is the edge set where each edge $e_{ij}$ represents the relationship between vertices $v_i$ and $v_j$. 
For each direction $\mathcal{G}^d$, let $\mathcal{A}_{\mathcal{G}^d(v_i)}$ and  $\mathcal{S}_{\mathcal{G}^d(v_i)}$ be the predecessor and successor sets of $v_i$.
Depending on the relationships between vertices in $\mathcal{G}^d$, all vertices are organized into a sequence and fed into the CRRN structure for modeling.
As illustrated in Fig. \ref{fig:forward}, the $i$-th CRRN unit takes an input of $\textbf{x}^{v_i}$ and the hidden state of all vertices in the predecessor set $\mathcal{A}_{G^d(v_i)}$.
The input $\textbf{x}^{v_i}$ can be computed as $\mathbf{x}^{v_i} = \text{vec}(v_i)$
where $\text{vec} \left( \cdot \right)$ is the vectorization operator. The following will detail the proposed CRRN architecture with two main tasks: (1) the contextual information embedding via sequence learning, and (2) the visual representation learning.

\subsubsection{Contextual Information Embedding}\label{subsec:CRRN_training}

Starting with one DAG $\mathcal{G}^d$ at vertex $v_i$, the contextual relationship between the vertex $v_i$ and its predecessor set $\mathcal{A}_{G^d(v_i)}$ is expressed as a non-linear function over the current input $x^{v_i}$ and the summation of hidden layers of all its predecessors in the context-based component as follows:
\begin{equation}
\begin{split}
\hat{\textbf{h}}_d^{v_i} &= f_{\text{CONTEXT}}(\mathbf{x}^{v_i}, \mathcal{A}_{\mathcal{G}^d(v_i)}; \theta_1) \\
& =  \phi(\mathbf{U}\mathbf{x}^{v_i} + \sum_{v_j \in \mathcal{A}_{\mathcal{G}^d(v_i)}}{\mathbf{W}\mathbf{h}_d^{v_j}} + \mathbf{b})
\end{split}
\label{eq:h_hat}
\end{equation}
where $\textbf{h}_d^{v_j}$ is the hidden state of vertex $v_j$ belonging to predecessor set $\mathcal{A}_{\mathcal{G}^d(v_i)}$ of vertex $v_i$ in the DAG $\mathcal{G}^d$, $\hat{\textbf{h}}_d^{v_i}$ is the intermediate hidden state of vertex $v_i$ and $\theta_1=\{\mathbf{U}, \mathbf{W}, \mathbf{V}, \mathbf{b}\}$ is the parameters of context-based component representing the connection weights of input-to-hidden, predecessor-to-hidden and hidden-to-output; and the hidden bias, respectively. $\phi \left(\cdot\right)$ is the activation function, i.e. ReLU function.

\subsubsection{Visual representation learning}

To further extract powerful visual representation as well as address the vanishing problem during modeling, we employ the visual-based component with residual learning technique. Given the intermediate hidden state $\hat{\textbf{h}}_d^{v_i}$ of the vertex $v_i$ provided by the context-based component,
The representation $\textbf{h}_d^{v_i}$ of each vertex $v_i$  is computed as in Eqn. \eqref{eq:h}.
\begin{equation}
\begin{split}
  \textbf{h}_d^{v_i} & = f_{\text{VISUAL}}(\hat{\textbf{h}}_d^{v_i}; \theta_2)\\
  & = \phi(\mathcal{F}(\hat{\textbf{h}}_d^{v_i}) + \mathcal{H}(\hat{\textbf{h}}_d^{v_i }))
\end{split}
  \label{eq:h}
\end{equation}
where 
$\mathcal{F}$ denotes a residual function consisting of a two-layer convolutional network in residual-network style, i.e equipped with an residual learning connection as shown in Fig. \ref{fig:forward}. This network is a stack of two convolutional layers, i.e. alternating convolution, batch normalization and ReLU operations. Meanwhile, $\mathcal{H}$ is defined as an identity mapping, where $\mathcal{H}(\hat{\textbf{h}}_d^{v_i }) = \hat{\textbf{h}}_d^{v_i }$. $\theta_2$ represents the parameters of the two-layer convolutional network.

The final output from the four DAGs are then combined using the following Eqn.\ref{eq:final_o}
\begin{equation}
\begin{split}
\textbf{o}^{v_i} & =f(\mathbf{x}^{v_i}, \mathcal{A}_{\mathcal{G}(v_i)};\theta_1, \theta_2)\\
& = \sum_{d=1}^4 \mathbf{V}\hat{\mathbf{h}_d}^{v_i} + \mathbf{b}_o
\end{split}
\label{eq:final_o}
\end{equation}

where $\textbf{V}$ is the hidden-to-output weight matrix, $\mathbf{b}_o$ is output bias.  
The CRRN is then optimized to minimize the negative log-likelihood over the training data as follows.
\begin{equation}
\theta_1^*, \theta_2^* = \arg \min_{\theta_1, \theta_2} L(\theta_1,\theta_1)
\end{equation} 
\begin{eqnarray}
\begin{split}
L(\theta_1,\theta_1) = -\frac{1}{n}\sum_{v_i \in \mathcal{G}}\sum_j\log p(l^{v_i}_j|\mathbf{x}^{v_i};\theta_1, \theta_2)\\
p(l^{v_i}_j|\mathbf{x}^{v_i};\theta_1, \theta_2) = \frac{e^{f_{l^{v_i}_j}(\mathbf{x}^{v_i}, \mathcal{A}_{\mathcal{G}(v_i)};\theta_1, \theta_2)}}{\sum_{c=1}^C e^{f_{c}(\mathbf{x}^{v_i}, \mathcal{A}_{\mathcal{G}(v_i)};\theta_1, \theta_2)}}
\end{split}
\end{eqnarray}

where  $C$ is the number of classes; $n$ is the number of images; $l^{v_i}_j$ is the correct label of $j$-th pixel in block $v_i$; and $f_{l^{v_i}_j}(\mathbf{x}^{v_i}, \mathcal{A}_{\mathcal{G}(v_i)};\theta_1, \theta_2) = o_j^{v_i}$.

\subsection{Model Learning} \label{sec:Model_Learning}
\begin{figure}[t]
	\centering \includegraphics[width=0.8\columnwidth]{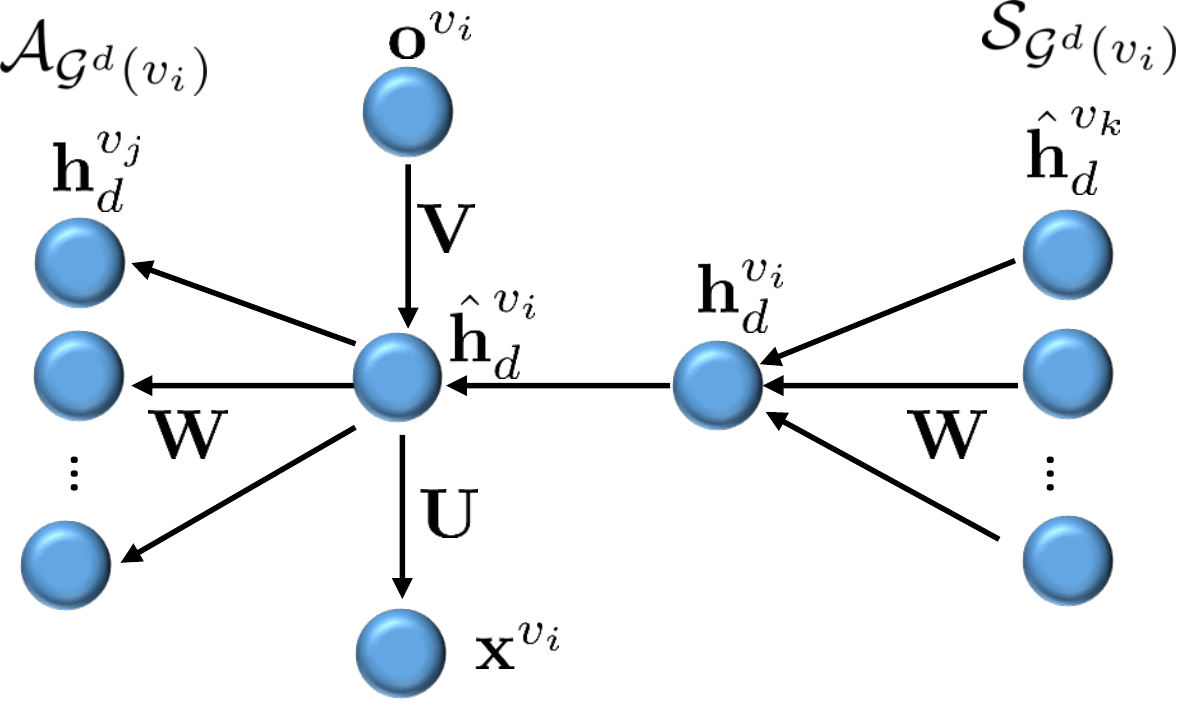}
		\caption{The backward procedure of CRRN at vertex $\textbf{v}_i$}
	\label{fig:backward}
\end{figure}

The optimal parameters can be obtained with the Stochastic Gradient Descent (SGD) algorithm given by
\begin{equation}
\theta_1 \leftarrow \theta_1 - \lambda \frac{\partial L}{\partial \theta_1}; \theta_2 \leftarrow \theta_2 - \lambda \frac{\partial L}{\partial \theta_2}
\end{equation}
where $\lambda$ denotes the learning rate.
\label{subsec:backward}
The derivatives are computed in the backward pass procedure is processed in the reverse order of forward propagation sequence as illustrated in Fig.\ref{fig:backward}. Instead of looking at predecessor $\mathcal{A}_{\mathcal{G}^d(v_i)}$ in the forward pass, we are now taking a look at successor $\mathcal{S}_{\mathcal{G}^d(v_i)}$.  
It is clear that the backpropagation error at the intermediate hidden layer $d\hat{\textbf{h}}_d^{v_i}$ comes from two sources: one from output $\frac{\partial \textbf{o}_d^{v_i}}{\partial \hat{\textbf{h}}_d^{v_i}}$ and the other one from hidden layer $\frac{\partial \textbf{h}_d^{v_i}}{\partial \hat{\textbf{h}}_d^{v_i}}$ whereas the backpropagation error at hidden layer $d{\textbf{h}}_d^{v_i}$ comes from its successor set $\mathcal{S}_{\mathcal{G}^d(v_i)}$. i.e, $\sum_{v_k \in \mathcal{S}_{\mathcal{G}^d(v_i)}}{\frac{\partial \textbf{o}_d^{v_k}}{\partial \hat{\textbf{h}}_d^{v_k}}\frac{\partial \hat{\textbf{h}}_d^{v_k}}{\partial \textbf{h}_d^{v_i}}}$. 
\normalsize
To make it simple and general, we would like to express the parameters in a form without subscript $_d$. For example we will use $\mathbf{h}^{v_i}$ instead of $\mathbf{h}_d^{v_i}$.
The derivative with respect to the model parameters \{$\mathbf{o}^{v_i}$, $\hat{\mathbf{h}}^{v_i}$, $\mathbf{h}^{v_i}$, $\mathbf{V}$, $\mathbf{U}$, $\mathbf{W}$\} can be computed as follows. 
\begin{eqnarray}
\begin{split}
d\mathbf{o}^{v_i} &= \frac{\partial L}{\partial \mathbf{o}^{v_i}}\\
d \hat{\textbf{h}}^{v_i} &= \frac{\partial L}{\partial \textbf{o}^{v_i}}\frac{\partial \textbf{o}^{v_i}}{\partial \hat{\textbf{h}}^{v_i}} + \left(\frac{\partial \textbf{h}^{v_i}}{\partial \hat{\textbf{h}}^{v_i}}\right)^T d \textbf{h}^{v_i}\\ 
& = \textbf{V}^T d  \textbf{o}^{v_i}  + \left(\frac{\partial \textbf{h}^{v_i}}{\partial \hat{\textbf{h}}^{v_i}}\right)^T d \textbf{h}^{v_i}\\
d\textbf{h}^{v_i} &=\sum_{v_k \in \mathcal{S}_{\mathcal{G}^d(v_i)}}{\textbf{W}^T \frac{\partial L}{\partial \hat{\textbf{h}}^{v_k}}\frac{\partial \hat{\textbf{h}}^{v_k}}{\partial \textbf{h}^{v_i}}} \circ \phi'(\hat{\textbf{h}}^{v_i})\\
& = \sum_{v_k \in \mathcal{S}_{\mathcal{G}^d(v_i)}}{\textbf{W}^T d\hat{\textbf{h}}^{v_k}} \circ \phi'(\hat{\textbf{h}}^{v_i})\\
\nabla\textbf{V} &= \frac{\partial L}{\partial \textbf{o}^{v_i}}\frac{\partial \textbf{o}^{v_i}}{\partial \textbf{V}} = d  \textbf{o}^{v_i} \left(\hat{\textbf{h}}^{v_i}\right)^T\\
\nabla\textbf{U} &= \frac{\partial L}{\partial \textbf{o}^{v_i}}\frac{\partial \textbf{o}^{v_i}}{\partial \hat{\textbf{h}}^{v_i}} \frac{\partial \hat{\textbf{h}}^{v_i}}{\partial\textbf{U}}= d  \hat{\textbf{h}}^{v_i} \circ \phi'(\hat{\textbf{h}}^{v_i})\left(\textbf{x}^{v_i}\right)^T\\
 \nabla\textbf{W} &= \frac{\partial L}{\partial \textbf{o}^{v_i}}\frac{\partial \textbf{o}^{v_i}}{\partial \hat{\textbf{h}}^{v_i}} \frac{\partial \hat{\textbf{h}}^{v_i}}{\partial \textbf{W}}\\
 & =  \sum_{v_j \in \mathcal{S}_\mathcal{G}^d(v_i)}{d\hat{\textbf{h}}^{v_j}\circ \phi'(\hat{\textbf{h}}^{v_j})\left(\hat{\textbf{h}}^{v_i}\right)^T }
\end{split}
\label{eq:derive_params}
\end{eqnarray}

where $\circ$  represents the Hadamard product.

\subsection{Inference}
\label{subsec:CRRN_inference}
Given a testing image $\mathbf{I}$, we first divide $\mathbf{I}$ into $N$ blocks $v_i$, $i = 1 ... N$. These blocks are then fed into four DAGs. The inference process of each pixel $j$ in block $v_i$ can be performed by finding the class label that maximizes the conditional probability given by
\begin{equation}
	l^{v_i*}_j =\arg \max_c p(c|\mathbf{x}^{v_i};\theta_1, \theta_2)
\end{equation}
Finally, the prediction maps of all $v_i$ are concatenated based on the location of block $v_i$ in $\mathbf{I}$ for the final prediction map. Fig. \ref{fig:CRRN_architecture} illustrates the inference process of CRRN at one direction (southeast) as an example.

\section{Experimental Results}
In this section, we evaluate the efficiency of our proposed CRRN using four challenging and popular scene image labeling datasets, i.e. SiftFlow \cite{liu2009nonparametric}, CamVid \cite{brostow2008segmentation}, Stanford Background \cite{gould2009decomposing} and SUN \cite{xiao2010sun}.

\subsection{Datasets and Measurements}
\textbf{SiftFlow Dataset}
\cite{liu2009nonparametric} contains 2,688 images captured from 8 typical outdoor scenes, i.e. coast, forest, highway, inside city, mountain, open country, street, tall building. Each image is $256 \times 256$ and labeled with 33 semantic classes (ignore the background). To run the experiment, the dataset is separated into 2 subsets corresponding to 2,488 images for training and the rest for testing as in \cite{shuai2015dag}. 

\textbf{CamVid Dataset}
\cite{brostow2008segmentation} is a road scene dataset which contains 701 high-resolution images of 4 driving videos captured at both day and dusk modes \cite{CamVid_2009}. Each image is 960x720 and is labeled with 32 semantic classes. We follow the usual split protocol as given in \cite{tighe2013finding} with 468 images for training and the rest for testing. 

\textbf{Stanford Background Dataset} 
\cite{gould2009decomposing} contains 715 images annotated with 8 semantic classes. The dataset is randomly partitioned into 80\% (572 images) for training and the rest (143 images) for testing with 5-fold cross validation. 

\textbf{SUN Dataset} 
\cite{xiao2010sun} contains 16,873 images annotated with 3,819 semantic object categories. However, the numbers of images per category are highly unbalanced. Therefore, we choose the most common 33 categories for evaluation. 
The chosen dataset with these categories contains 6,433 images and is randomly partitioned into two subsets corresponding to 5,798 images for training and the rest for testing.

\textbf{Measurements} It has become common practice to report results using two metrics, namely, per-pixel accuracy (PA) and average per-class accuracy (CA). The first metric, i.e. $\text{PA} = \sum_i n_{ii}/\sum_i t_i$, is defined as the fraction of the number of pixels classified rightly over the number of pixels to be classified in total whereas the latter metric, i.e. $\text{CA} =(1/C)\sum_i (n_{ii}/t_i)$, is defined as the average of per-pixel accuracy of all the classes existing in the dataset. 
$n_{ij}$ is the number of pixels of class $i$ that were predicted to be class $j$, $C$ is total number of classes and $t_i = \sum_j n_{ij}$ is the total number of pixels of class $i$.

\begin{figure}[t]
	\centering \includegraphics[width=0.75\columnwidth]{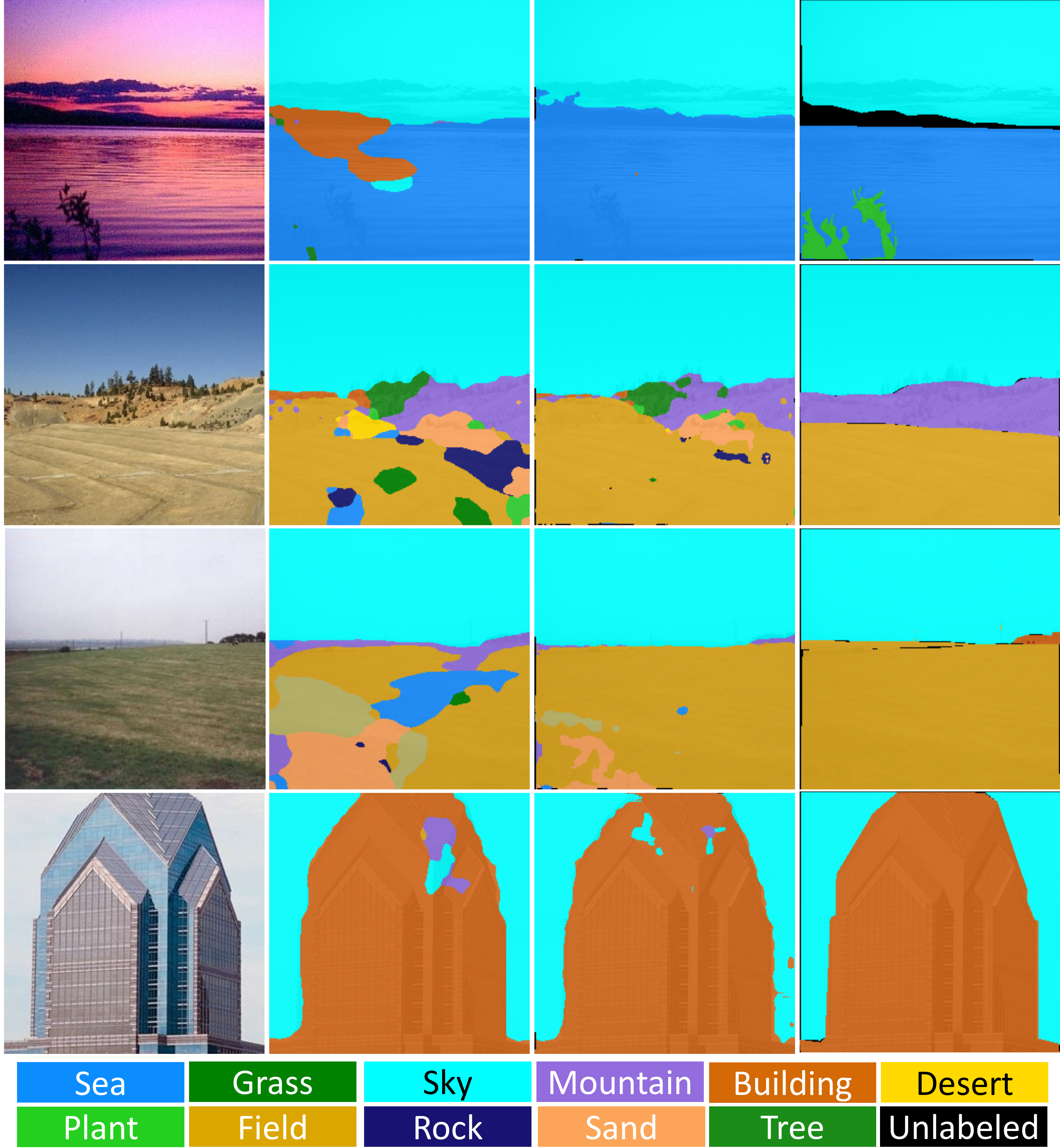}
		\caption{Comparison between our CRRN and \cite{shuai2015dag} in term of contextual dependencies learning. In each row, there are four images: input image (first column), the prediction labeling map of \cite{shuai2015dag} (second column), the contextual labeling map extracted from our CRRN (third column) and the ground truth labels. \textbf{(Best viewed in color.)}}
	\label{fig:sample_Contextual_Learning}
\end{figure}

\begin{figure*}[!t]
	\centering \includegraphics[width=1.75\columnwidth]{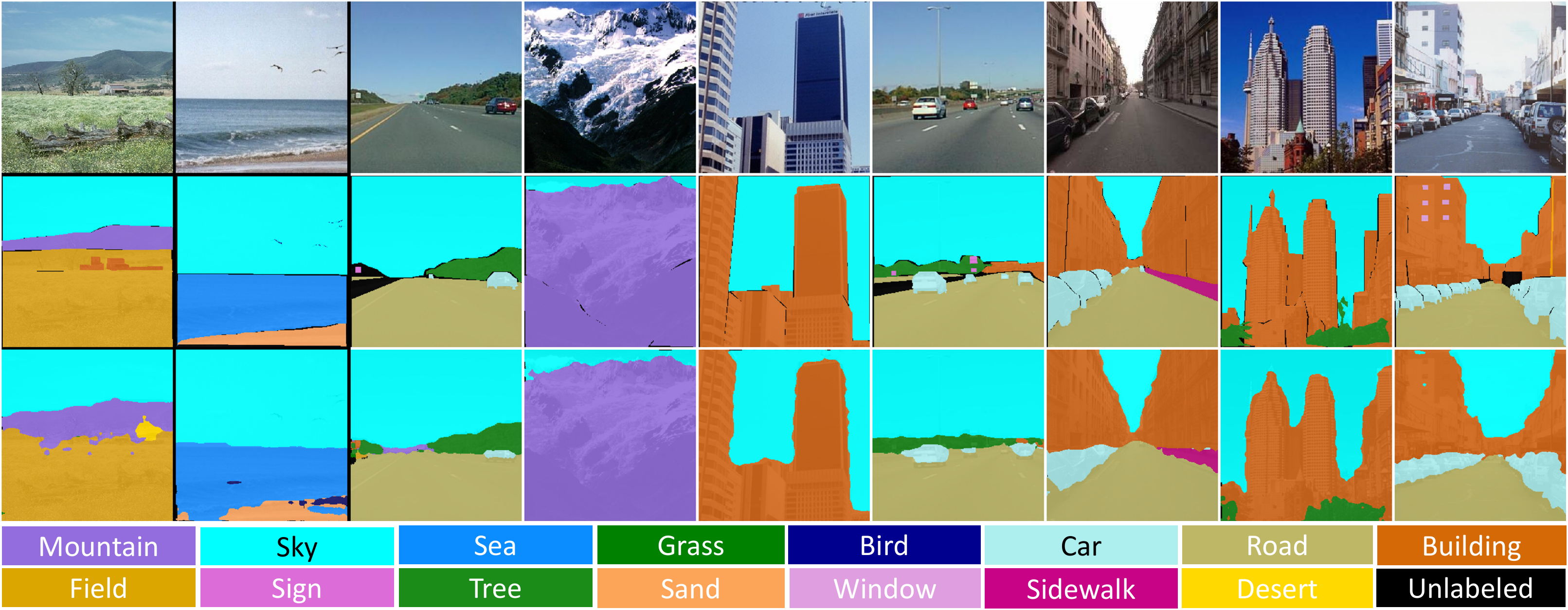}
		\caption{Examples of labeling results on SiftFlow dataset. Each column has three images: the input image (first row), its ground truth labels (second row) and our CRRN prediction labels (third row), respectively. \textbf{(Best viewed in color.)}}
	\label{fig:sample_siftflow}
\end{figure*} 

\subsection{Implementation Details}

In this section, the implementation of our proposed model is discussed in details. 
The model is implemented in TensorFlow environment and runs in a machine of Core i7-6700 @3.4GHz CPU, 64.00 GB RAM and a single NVIDIA GTX Titan X GPU.
In order to train the CRRN, each image in the training data is first divided into 64 (= $8 \times 8$) blocks. Next, these blocks are reorganized into four sequences based on their relationships in the four DAGs.
Each block is vectorized and used as the input for the CRRN unit. Our CRRN architecture is then employed to simultaneously model the contextual dependencies between blocks in different directions as well as extract their compact and rich representation.
The dimensionality of the hidden layer is set to 256 in the recurrent component and reshaped to $16 \times 16$ before going through the residual learning component. From our preliminary results, hidden layers with smaller number of hidden units are not powerful enough to capture both contextual information and visual representation in four DCGs. 
The optimal parameter values of the whole network are obtained via a stochastic gradient descent procedure with forward and backward passes. The learning rate is initialized to be $10^{- 3}$, and decays with the rate of 0.95 after 30 epochs.
The whole model takes about 12.7 hours to train. For inference process, CRRN takes 0.058 second/image.
\begin{table}[!t]
\centering
\caption{Quantitative results and comparisons against \textit{non-fine-tuned models} \textbf{2D-LSTM} \cite{byeon2015scene}, \textbf{Recurrent CNN} \cite{pinheiro2014recurrent}, \textbf{Multi-scale Convnet}  \cite{farabet2013learning}, \textbf{Multi-CNN - rCPN} \cite{Sharma2014_recursive}, S\textbf{cene Graph Structure} \cite{Souly_2016_sparse}, \textbf{CNN-65-DAG-RNN} \cite{shuai2015dag}, \textbf{Sample\&Filter + MRF} \cite{Najafi2015_nonparametric}, \textbf{Multi-scale RCNN} \cite{Liang2015_recurrent} on Siftflow dataset.}
\label{tab:res_siftflow}
\begin{tabular}{lcc}
\toprule[\headrulewidth]
\textbf{METHOD} & \textbf{PA} & \textbf{CA} \\ \midrule
2D-LSTM      & 70.1\%                   & 20.9\%  \\ 
Recurrent CNN  & 77.7\%                    & 29.8\%   \\ 
Multi-scale Convnet   & 78.5\%                    & 29.4\%     \\ 
Multi-CNN - rCPN   & 79.6\%                    & 33.6\%  \\ 
Scene Graph Structure    & 80.6\%                    & 45.8\%            \\ 
CNN-65-DAG-RNN     & 81.1\%                    & 48.1\%          \\ 
Sample\&Filter + MRF   & 83.1\%                    & 44.3\%     \\ 
Multi-scale RCNN   & 83.5\%                    & 35.8\%         \\ 
\textbf{Our CRRN}                 &            \textbf{84.7\%}           &       \textbf{61.0\%}             \\
\bottomrule[\headrulewidth]
\end{tabular}
\end{table}

\begin{table}[!t]
\centering
\caption{Quantitative results and comparisons against \textit{non-fine-tuned models} \textbf{Neural Decision Forest}s\cite{rota2014neural}, \textbf{SVM+MRF}\cite{tighe2013finding} on CamVid dataset.}
\label{tab:res_Camvid}
\begin{tabular}{lcc}
\toprule[\headrulewidth]
\textbf{METHOD} & \textbf{PA} & \textbf{CA} \\ \midrule
Neural Decision Forests & 82.1\% & 56.1\% \\
SVM+MRF & 83.9\% & \textbf{62.5\%} \\
\textbf{Our CRRN}                  &        \textbf{84.4\%} &       54.8\%         \\
\bottomrule[\headrulewidth]
\end{tabular}
\end{table}

\subsection{Scene Labeling Results}
In order to make a fair comparison between our proposed CRRN and other state-of-the-art models, we divide the models into two groups: the ones trained using the benchmark datasets only; and the ones fine-tuned from other models which are trained over the large-scale datasets (i.e. ImageNet) or made used of deep network (i.e. VGG-verydeep-16) as their feature extractor. \textit{Our CRRN falls into the first group where no pre-trained model is used.}
\begin{table}[t]
\centering
\caption{Quantitative results and comparisons against \textit{non-fine-tuned models} \textbf{2D-LSTM} \cite{byeon2015scene}, \textbf{Recurrent CNN} \cite{pinheiro2014recurrent}, \textbf{Multi-scale Convnet}  \cite{farabet2013learning}, \textbf{Multi-scale RCNN} \cite{Liang2015_recurrent}, \textbf{Scene Graph Structure} \cite{Souly_2016_sparse} on  Stanford-background Dataset}
\label{tab:res_Stanford}
\begin{tabular}{lcc}
\toprule[\headrulewidth]
\textbf{METHOD} & \textbf{PA} & \textbf{CA} \\ \midrule
2D-LSTM     & 78.6\%                  & 68.8\%         \\ 
Recurrent CNN  & 80.2\%                    & 69.9\%   \\ 
Multi-scale Convnet    & 81.4\%                    & 76.0\%       \\ 
Multi-scale RCNN    & 83.1\%                    & 74.8\%          \\ 
Scene Graph Structure    & 84.6\%                    & \textbf{77.3\%}         \\ 

\textbf{Our CRRN}                  &        \textbf{85.23\%}                 &          75.2\%               \\ 
\bottomrule[\headrulewidth]
\end{tabular}
\end{table}

Comparing to the methods in the first group, the quantitative results of our approach with three benchmark datasets namely, Siftflow, Camvid, Stanford and SUN are reported in Tables \ref{tab:res_siftflow}, \ref{tab:res_Camvid}, \ref{tab:res_Stanford} and \ref{tab:compare_LMSUN}, respectively. The empirical results on three datasets show that our performance on both PA and CA scores are higher than state-of-the-art methods on larger dataset while giving quite competitive results on the smaller dataset. On larger Siftflow  when we have big enough data for training, our proposed CRNN outperforms all other models in terms of both PA and CA scores. Our CA is $61.0\%$ and PA is $84.0\%$ compared to the next highest CA score of $48.1\%$ \cite{shuai2015dag} and PA score of $83.5\%$ \cite{Liang2015_recurrent} as shown in Tables \ref{tab:res_siftflow}. On the small dataset, take Camvid as an instance, when we just have about 468 images for training, our PA is still about $0.5\%$ higher than the state-of-the-art method \cite{tighe2013finding}. On larger scale dataset such as SUN, our CRRN also achieves 1.41\% higher than FCNN \cite{long2015fully} in term of PA score.

\begin{figure*}[!t]
	\centering \includegraphics[width=1.75\columnwidth]{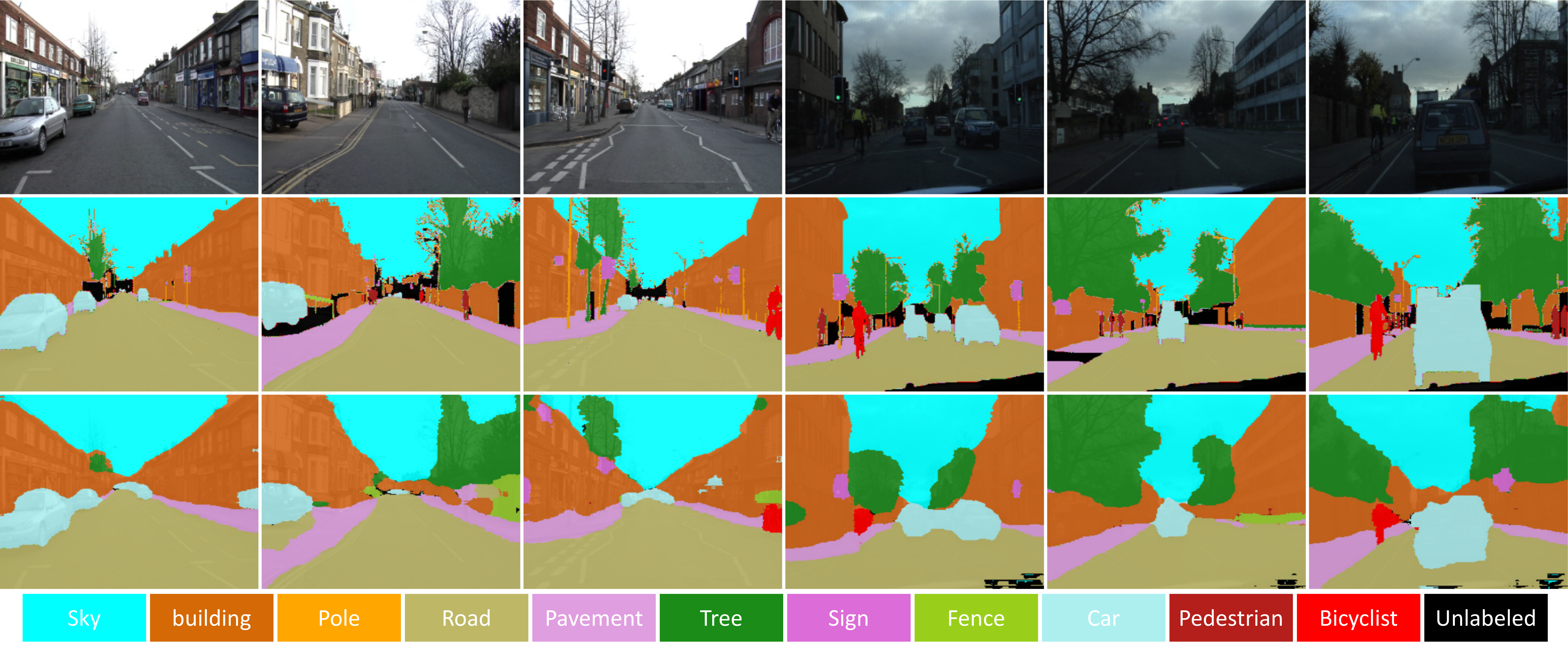}
		\caption{Examples of labeling results on CamVid dataset. Each column has three images: the input image (first row), its ground truth labels (second row) and our CRRN prediction labels (third row), respectively. \textbf{(Best viewed in color.)}}
	\label{fig:sample_Cam}
\end{figure*}
\begin{figure*}[!t]
	\centering \includegraphics[width=1.75\columnwidth]{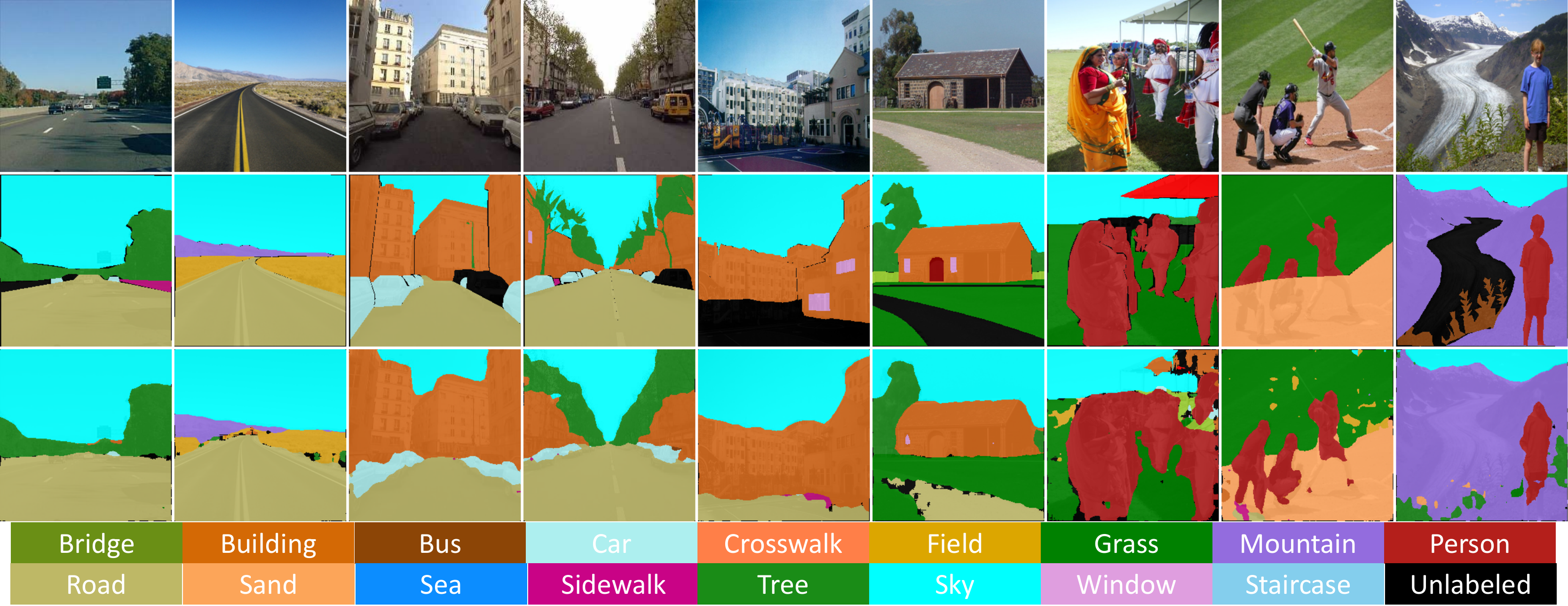}
		\caption{Examples of labeling results on SUN dataset. Each column has three images: the input image (first row), its ground truth labels (second row) and our CRRN prediction labels (third row), respectively. \textbf{(Best viewed in color.)}}
	\label{fig:sample_SUN}
\end{figure*}
\begin{table}[t]
\centering
\caption{Quantitative results and comparisons against \textit{non-fine-tuned models} \textbf{CNN-65-DAG-RNN}  \cite{shuai2015dag}, \textbf{FCNN} \cite{long2015fully} on SUN dataset.}
\label{tab:compare_LMSUN}
\begin{tabular}{lcc}
\toprule[\headrulewidth]
\textbf{METHOD} & \textbf{PA} & \textbf{CA} \\ \midrule
CNN-65-DAG-RNN  &	71.51$\%$	&	54.57$\%$ \\
FCNN & 77.2$\%$ & \textbf{62.03}$\%$ \\
\textbf{Our CRRN} & \textbf{78.61}$\%$ & 59.9$\%$ \\
\bottomrule[\headrulewidth]
\end{tabular}
\end{table}
\begin{table}[!t]
\centering
\caption{Quantitative results and comparisons against \textbf{CNN - Global Context} \cite{shuai2015integrating}, \textbf{FCNN} \cite{long2015fully}, \textbf{VGG-conv5-DAG-RNN} \cite{shuai2015dag} with \textit{fine-tuned models} on Siftflow dataset.}
\label{tab:compare_finetune_siftflow}
\begin{tabular}{lcc}
\toprule[\headrulewidth]
\textbf{METHOD} & \textbf{PA} & \textbf{CA} \\ \midrule
CNN - Global Context    & 80.1\%                    & 39.7\%        \\
FCNN  & 85.2\% & 51.7\% \\
VGG-conv5-DAG-RNN     & \textbf{85.3\%}                    & 55.7\%                 \\ 
\hline
\textbf{Our CRRN}                 &            84.7\%           &       \textbf{61.0\%}               \\
\bottomrule[\headrulewidth]
\end{tabular}
\end{table}
Figure \ref{fig:sample_Contextual_Learning} illustrates the advantages of our CRRN in term of modeling the contextual dependency presented in the image. Comparing to \cite{shuai2015dag}, besides the local consistency between neighborhood regions, their semantic coherence is better enhanced in our CRRN. For example, after training, our CRRN can capture the contextual knowledge such as the `building' is not likely to appear in the 'sea' (i.e. the first case) or the 'desert' is not usually covered by the 'field' (i.e. the second case). As a result, while \cite{shuai2015dag} still has the problem of misclassification in its predictions, smoother and better labeling maps can be produced by our model.
Our qualitative results on the datasets are further demonstrated in Figures \ref{fig:sample_siftflow}, \ref{fig:sample_Cam} and \ref{fig:sample_SUN}.  As visualization, there exist many false positive cases where pixels are classified as unlabeled while they truly belongs to labeled classes. In such cases, our CRRN performs well to category the false positive pixels. Thank to the ability of simultaneously learning rich visual representation and modeling the context information, our model has better memory of contextual dependency and gives higher discriminative descriptors for each local patch

Furthermore, we also compare our CRRN with other fine-tuned methods on SiftFlow as shown in Table \ref{tab:compare_finetune_siftflow}. In this group, most existing fine-tune methods \cite{shuai2015integrating,long2015fully,shuai2015dag} are trained on large-scale data and make use of powerful feature extractor while our CRRN is a non fine-tune model and trained on Siftflow only. From these results, one can see that our CRRN model archives the best CA score ($61.0\%$) compared to the state-of-the-art ($55.7\%$) \cite{shuai2015dag} while giving a competitive PA score. We believe that given enough data, our CRRN performance can be boosted and is competitive to these models. We leave this as future work.

\section{Conclusion}

This paper presents a novel Contextual Recurrent Residual Networks (CRRN) approach, which is able to simultaneously model the long-range context dependencies and learn rich visual representation. The proposed CRRN is designed as a fully end-to-end deep learning framework and is able to make use the advantages of both sequence modeling and residual learning techniques. Our CRRN network contains three parts corresponding to sequential input data, sequential output data and hidden CRRN unit. Each hidden CRRN unit has two main components: context-based component to model the context dependencies and visual-based component to learn the visual representation. Our proposed end-to-end CRRN is completely trained from scratch, without using any pre-trained models in contrast to most existing methods usually fine-tuned from the state-of-the-art pre-trained models, e.g. VGG-16, ResNet, etc. The experiments are conducted on four challenging scene labeling datasets, i.e. SiftFlow, CamVid, Stanford background and SUN datasets, and compared against various state-of-the-art scene labeling methods.

\bibliography{egbib}
\bibliographystyle{icml}

\end{document}